\UseRawInputEncoding 

\documentclass[runningheads]{llncs}

\usepackage[utf8]{inputenc}

\usepackage{cite}
\usepackage{color} 
\usepackage{graphicx}
\usepackage{array}
\newcolumntype{P}[1]{>{\centering\arraybackslash}p{#1}}
\usepackage{multirow}
\usepackage{amssymb}
\usepackage{amsmath}
\usepackage{relsize}
\usepackage{lipsum}
\usepackage[multiple]{footmisc}

\DeclareMathOperator{\readout}{READOUT}

\begin{document}
\title{Controversy Detection: a Text and Graph Neural Network Based Approach }
\titlerunning{Controversy Detection: a Text and Graph Neural Network Based Approach}


%


\author{Samy Benslimane\inst{1} \and
J\'erome Az\'e \inst{1} \and
Sandra Bringay \inst{1,2} \and \\
Maximilien Servajean \inst{1,2} \and
Caroline Mollevi \inst{3,4}}

\authorrunning{S. Benslimane et al.}

\institute{LIRMM UMR 5506, CNRS, University of Montpellier, Montpellier, France \and
AMIS, Paul Valery University, Montpellier, France,
\and Institut du Cancer Montpellier (ICM), Montpellier, France
\and IDESP, UMR Inserm - Université de Montpellier, Montpellier, France \\
\email{\{first.last\}@lirmm.fr, caroline.mollevi@icm.unicancer.fr}}
\maketitle


\begin{abstract}
Controversial content refers to any content that attracts both positive and negative feedback.
Its automatic identification, especially on social media, is a challenging task as it should be done on a large number of continuously evolving posts, covering a large variety of topics.
Most of the existing approaches rely on the graph structure of a topic-discussion and/or the content of messages. 
This paper proposes a controversy detection approach based on both graph structure of a discussion and text features.
Our proposed approach relies on Graph Neural Network (\textsc{gnn}) to encode the graph representation (including its texts) in an embedding vector before performing a graph classification task. The latter will classify the post as controversial or not. 
Two controversy detection strategies are proposed. The first one is based on a hierarchical graph representation learning. Graph user nodes are embedded hierarchically and iteratively to compute the whole graph embedding vector.
The second one is based on the attention mechanism, which allows each user node to give more or less importance to its neighbors when computing node embeddings.
We conduct experiments to evaluate our approach using different real-world datasets.
Conducted  experiments  show the positive impact of combining textual features and structural information in terms of performance.


\end{abstract}

\section{Introduction} 


The availability of large amount of data sources and the emergence of various social networks including Twitter and Reddit, increased the social connectivity of people. This allowed them to easily express, propagate, share and dispute opinions and gives us a great opportunity to study and understand social phenomena like controversial topics.
Expressed opinions through posts and articles often trigger fierce and sometimes endless debates, and frequently cause a controversy.
A controversial content can  simply be defined as any content that attracts both positive and negative feedback~\cite{Hessel2019Controversy}. Indeed, polarization stigmatizes user's behavior in presence of controversial topics~\cite{Beelen2017NewsControversy, Garimella2018ReducingControversy, Garimella2017Quantifying}. 

Automatic controversy detection can be helpful. For instance, people can be warned by the existence of a controversy to add some nuance to better understand some issues. Objective information could  also be brought to people to prevent misinformation or hateful discussions~\cite{Garimella2018ReducingControversy}. Detecting a content as non controversial is also helpful as it shows that people agree on a given issue. 

Automatic controversy identification is difficult and constitutes a challenging task as it should be done on a large number of continuously evolving posts covering a wide range of topics.
 This difficulty is increased by the fact that controversy is sometimes time-aware (what is controversial today was not necessary controversial in the past) and community-aware (what is controversial in a community is not necessary controversial in another community)~\cite{Jang2017ControversiInPopulation}.
\newline
Controversy analysis on web pages or articles is usually based on features extracted  from the content~\cite{Sznajder2019Controversy, Jang2016ProbabilisticAT}. 
However, on social media, the interaction between people is widely used to detect controversy.
These interactions include social relation (retweet and follow on Twitter, comment on Reddit)~\cite{Garimella2017Quantifying, Morales2015PoliticalPolarization}  and citation relation on Wikipedia~\cite{Jang2016ImproveAutomatedCD}.

In this paper, we focus on detecting controversy on the social media Reddit, even if any other social media (Twitter, Facebook, etc.) can  also be used after making few adaptations of the graph building stage. 
The originality of our approach firstly resides on using very recent state-of-the-art \textsc{gnn} methods to embed user nodes in a low d-dimensional space and take into account structural information. 
Initial text representations of a user are learnt from their messages on a current post and used as input of our \textsc{gnn}-based approach. To the best of our knowledge, only Zhong et al. started to use \textsc{gnn}s,
building their post-level controversy detection method by directly exploiting the comment-tree structure~\cite{Zhong2020ControversyGCN}.
Our work, on the contrary, exploits the user's interaction graph built from the comment-tree structure and 
compare multiple state-of-the art (\textsc{gnn}s) to combine both
structural and content information.


To detect controversial posts, we propose a \textsc{gnn}-based approach which
consists of the following main contributions:
\begin{itemize}

\item  \textsc{gnn} architecture-based controversy detection. Our controversy detection is based on a graph classification.
We propose two strategies to embed the whole graph structure.
The first strategy aims to exploit hierarchical structure that may exist in the user graph structure.
Graph information are aggregated across the edges iteratively and in a hierarchical way. In our work, we rely on the \textsc{diffpool} approach which encodes the whole graph by stacking several pooling layers~\cite{Ying2018Diffpool}. 
The second strategy is based on an attention mechanism. It aims to allow each user node to
judge which user neighbor is more or less important than the others, during the node
embedding process, according to the structure and features of the graph.

\item Experimental study. We conduct experiments on real-world datasets to evaluate the proposed 
\textsc{gnn}-based approach only using structural information. 
We show that our approach gets good performance compared to our baseline.

\item Textual features. We show that incorporating initial textual representation of users can improve the performance of our approach.

\end{itemize}
The rest of this paper is organized as follows. 
Section 2 provides some related work. 
Section 3 presents an overview of our approach to automatically detect controversy on social media.
Its different stages are described and formalized.
Section 4 presents the performance experiments and discusses the obtained results.
Section 5 concludes the paper and highlights some future work.



\section{Related Work} 
Works on controversy analysis can be classified in three groups: content-based, structure-based and hybrid approaches.

\textbf{Content-based.} Early methods to detect controversy are mainly based on textual features, and only focus on language semantic, supposing that immediate textual context of concept can be highly indicative \cite{Sznajder2019Controversy} or that text-content can be used as a tool for detecting controversial topic/post.
Several studies focus on the web controversy, thanks to sources like Wikipedia, where pages can automatically be labelled as controversial, using "edit-wars" \footnote{multiple editors on a Wikipedia concept exchanging opposing opinions.} and relations/citations between pages.
In~\cite{Sznajder2019Controversy},  an approach to measure how controversial a concept is on Wikipedia pages is proposed.
Instead of relying on Wikipedia's metadata, authors argue that immediate textual context of a concept is strongly indicative of controversiality.
They represent articles via pre-trained word embeddings methods
and define three controversiality estimators based on the nearest neighbors, naive Bayes, and recurrent neural network respectively.
In~\cite{Dori2016Collective}, authors were interested in identifying whether a given content on a web page is controversial or not.
The collective controversy classification model is based on a nearest neighbor classifier that identifies an article according to the related Wikipedia articles.
The idea is that if related Wikipedia articles are controversial, it is likely that the article is also controversial.
%
Other studies focus on probabilistic approaches~\cite{Jang2016ImproveAutomatedCD, Jang2016ProbabilisticAT} to combine Wikipedia Controversy meta-data and features like the \textit{MCD} score\footnote{presence of certain words,  ferocity of "edit-wars", etc.}.
Articles, from web media, are also a huge source of information.

\textbf{Structure-based.}
Textual messages on social media are usually biased and meanings might be different depending on many factors, such as the culture or language of the communities, and therefore should be treated with precaution. 
When studying controversy on a user interaction basis, the structural information of those interactions become particularly relevant, especially on social media. 
Each social media has its own code. For example, Twitter has specific features, such as 'Retweet'
and 'Follow'.
In~\cite{Garimella2017Quantifying}, a user hybrid graph, combining follow and retweet edges, is built for a topic, which is defined by a set of hashtags. After partitioning the graph on two distinct communities, different methods to measure the controversy are checked, including a random-walk-based controversy measure (RWC).
In a similar study \cite{Garimella2018ReducingControversy}, they use the same graph to quantify and reduce controversy, by connecting opposing sides and creating bridges between communities for more exposure.
In \cite{Emamgholizadeh2020BRWForQC}, a similar approach is used suggesting that we can level user commitment at their community by looking at their relation. They propose a new method using Biased Random-Walk and adapt a new controversy measure to quantify.
A previous research focused on more exposed node boundaries, with statistical polarization measures to evaluate controversy \cite{Guerra2013comBoundariesPolarization}.
In \cite{Mendoza2020PolarizationGENE}, the importance of users and named entities involved in a discussion are highlighted. They generate a conditioned graph on named entities partitioned, and quantify controversy using a RWC (Random-Walk Controversy) score.
Even if structural features are widely covered and seem to be a strong asset, not covering text features appears to be a huge loss of relevant information. 

\textbf{Hybrid methods.}
Recent studies focus on combining both structural and content information to avoid losing valuable features.
On this condition, Social media appears to be the ideal source, with the multiplicity of user interactions.
In~\cite{Dezarate2020VocQuantifyingControversy}, authors extend the work of Garimella and al.\cite{Garimella2017Quantifying} and propose a vocabulary-based controversy detection.
Using the partitioned user graph, tweets of the two selected communities are grouped by users, pre-processed, concatenated and labeled by the community name of their corresponding user.
They constitute the dataset which is used to train the text representation model FastText.
The controversy score is finally computed by using the embedding of the central users.
In~\cite{Hessel2019Controversy}, authors demonstrate that mixing
structural features (number of comments, max depth/total comment ratio, average node depth, etc.) of post-comment tree of a Reddit discussion with textual features outputted by language models such as BERT \cite{Devlin2018BERT} can improve predictive performance of early controversy post-level detection.
With the same objective, a Graph Convolutional Network based approach is proposed in~\cite{Zhong2020ControversyGCN} by Zhong et al. It aims to integrate information extracted from the comment-tree structure as well as content of post and its comments. A parallel multi-task classifier is added on their model to disentangle topic-related and topic-unrelated features for inter-topic detection.
Even with good performance, this approach presents some limits. The comment-tree structure of a post prevents us from exploiting user interactions, and the use of inter-topics relation might interfere too much with the main detection task. However, to the best of our knowledge, this is the first study which focuses on \textsc{gnn} for controversy analysis.\\
We present in this paper a new hybrid controversy detection approach, based on user graph interaction and state-of-the-art \textsc{gnn}s to combine valuable textual and structure information.

\section{Graph Neural Network-based Controversy Detection Approach}\label{controversydetectionapproach} 

\begin{figure*}[!htbp]
	\centering
		\includegraphics[width=1\textwidth]{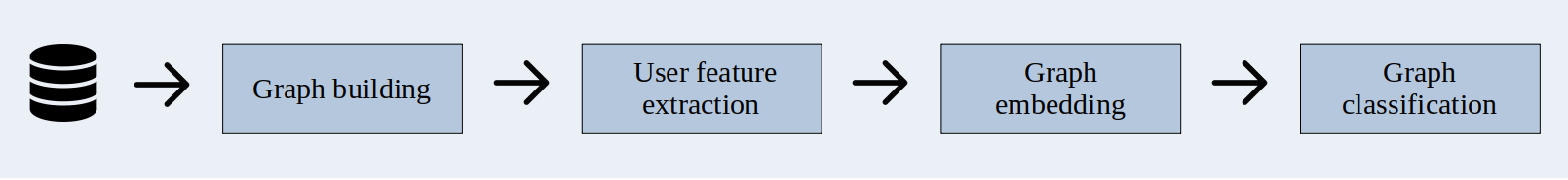}
	\caption{Overview of our controversy detection approach.}
	\label{fig:pipeline_apporach}
\end{figure*}

This section describes our  post-level controversy detection approach. It focuses on the  Reddit social media. Any other social media can be used with very few adaptations of the graph building stage. 
The main idea is to exploit both text content and user interactions by representing the Reddit discussion as a user graph and exploring advanced \textsc{gnn} embedding techniques.
 Figure~\ref{fig:pipeline_apporach} presents an overview of our approach. We divide our pipeline into four sequential stages: Graph Building, User Feature Extraction, Graph Embedding, and Graph Classification. 
 
The graph building stage represents data extracted from the Reddit social media as a user graph. 
We represent the initial comment-post tree as a graph where nodes represent users and edges correspond to the interaction that exists between users. 
Each node is represented by its own data (user-id, age, location, texts, etc.). 
The user feature extraction stage enriches graph nodes by adding textual embedded features. These features are computed by using state-of-the-art NLP techniques. This allows to better interpret the texts that users sent out.
The graph embedding stage computes the embedding of the whole graph.
Different advanced \textsc{gnn}-based graph representation learning techniques are used, namely \textsc{diffpool} \cite{Ying2018Diffpool}, \textsc{gcn} \cite{Kipf2017GCN}, 
and \textsc{gat-gc} \cite{Zhang2020GATGC}.
Finally, the graph classification stage predicts the binary label associated to the whole graph, that is to classify a post as controversial or not.


\begin{figure*}[!htbp]
	\centering
		\includegraphics[width=0.7\textwidth]{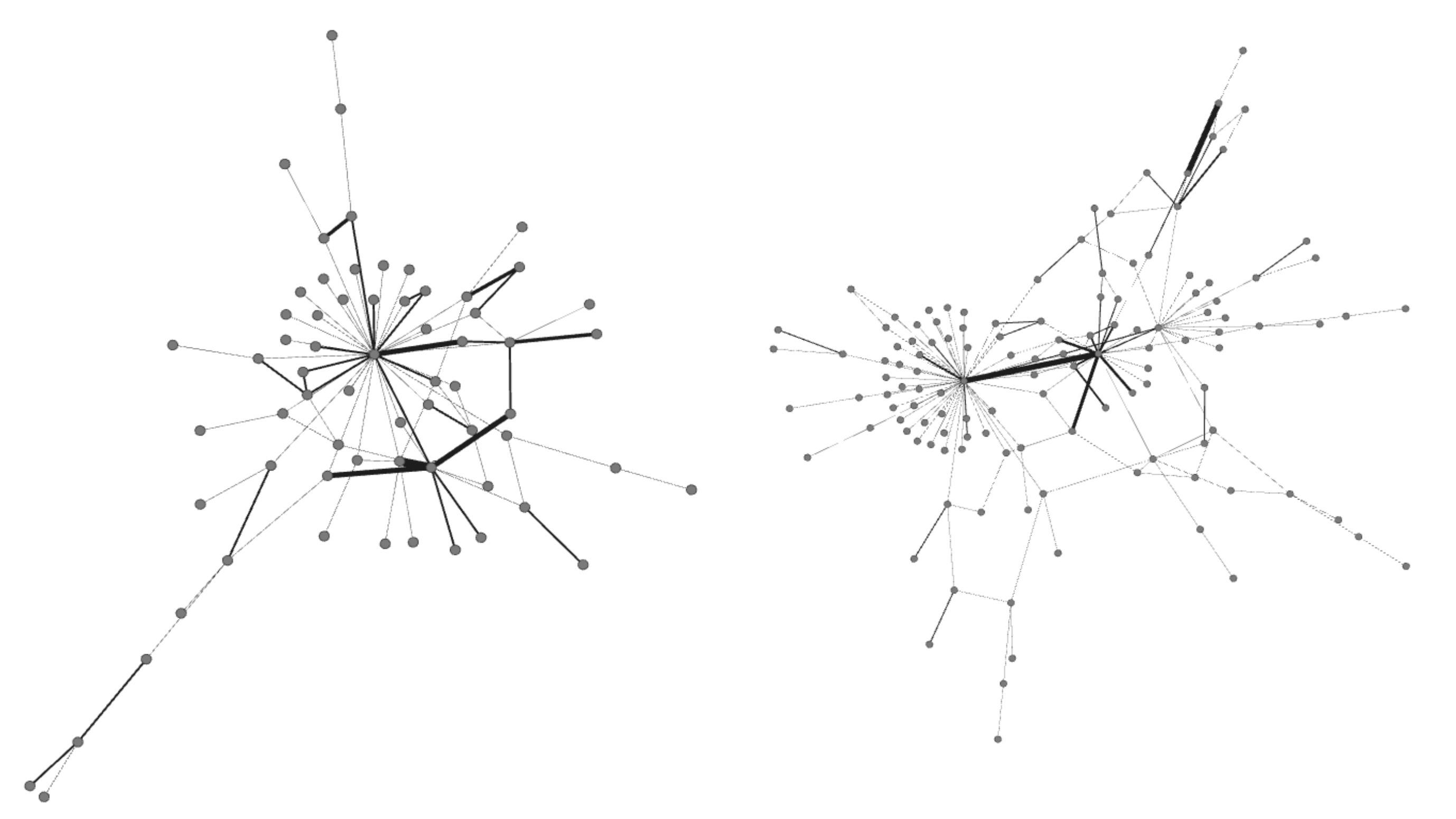}
	\caption{Left: User graph of a controversial post. Right: User graph of a non-controversial post. The more interaction there is between two nodes, the bolder the link is.}
	\label{fig:user_graph_reddit}
\end{figure*}

\subsection{Graph building} 

Existing controversy detection methods~\cite{Hessel2019Controversy, Zhong2020ControversyGCN} on Reddit use the classic comment-post tree representation as they mainly focus on the structure of the discussion.
However, many research works have established that user interaction can be helpful to extract different features on social media that can improve the controversy detection.
In this work, we adopt a graph representation of a discussion to highlight these user interactions.
Given a discussion on a post (thread) \(p\) extracted from a subreddit \(s\), we build an undirected graph where a node \(u_i\) represents a user involved in the discussion. An edge \((u_i,u_j)\) is created when a user \(u_j\) responds to the post $p$ or any comment posted by \(u_i\).
Figure~\ref{fig:user_graph_reddit} shows two user graphs of controversial and non-controversial posts respectively.

More formally, a post $p$ is represented as a graph \(G = (\mathcal{U}, \mathcal{E}, X)\)  where \(\mathcal{U}=\{u_1, u_2, ..., u_n\}\) denotes the user nodes, \(\mathcal{E}=\{(u_i,u_j)\}_{1\leq i,j\leq n} \)  denotes the edges of the graph, and \(X \in \mathbb{R}^{n\times e} \), \( e \) being the feature dimension,  denotes the user node features matrix. 
Each node corresponds to a unique user, and an edge between two nodes exists if there is interaction links between the corresponding users.
The computation of the matrix \(X\) is described in section~\ref{sec:node_feature_extraction}.

 

\subsection{User feature extraction}\label{sec:node_feature_extraction} 
In order to bring valuable information to the graph representation, user features are extracted from the posted texts by using advanced NLP techniques.
Recently, different NLP language models pre-trained on a large corpus have been proposed to improve the dynamic text representation, such as BERT \cite{Devlin2018BERT}.

The user features extraction is performed for each user as follows. Each message (post or comment) a user \(u_i \) posts is firstly cleaned (reddit tags and url link removed) and is then
embedded in an \(e\)-dimensional vector by using  a language model BERT~\footnote{more details on section \ref{sec:experiment}.}.
The embedded vectors obtained from the different messages posted by a user \(u_i \) are  aggregated to form the final
user features \( x_{u_i}\) as shown in equation~\ref{featuresaggregation}.


\begin{equation}
    x_{u_i} = \textsc{Aggregation} \bigg( [ x^0_{{u_i}}, x^1_{{u_i}}, ..., x^m_{{u_i}}] \bigg) 
    \label{featuresaggregation}
\end{equation}
where $x_{u_i}^j \in \mathbb{R}^{e}$ is the individual $e$-dimensional embedded vector computed from the $j^{th}$ message of user \(u_i\), and $m$ is the number of messages a user $u_i$ posted.
In this paper, the aggregation of the embedded text vectors is performed via the Max-pooling function, but any other aggregation function can be used. Features of each user $u_i$ is stacked on a matrix \(X \in \mathbb{R}^{n\times e} \).

The user graph \(G = (\mathcal{U}, \mathcal{E},  X) \) is now fully represented and includes node textual features. It will also be referenced by \((A,X)\) where \(A\) represents its adjacency matrix.

\subsection{Graph embedding}\label{sec:graph_embedding} 

The graph embedding stage aims to encode the whole user graph in a low-dimensional vector. This latter will fed
the graph classification stage to predict if a post is controversial or not.
Recently, different \textsc{gnn}-based approaches were proposed to adapt deep learning architectures to the graph structured data \cite{Kipf2017GCN, Velickovic2018GAT}.
The main idea is to consider each graph node as a computation node, and to learn classical neural network primitives that compute node embeddings.  


This stage relies on \textsc{gnn} architectures with the objective to exploit both user node features computed in the previous stage and the user graph structure of the Reddit discussion. The output is the embedding of the whole graph denoted by $z_{G}$. Learning individual node embeddings denoted by $z_{u_i}$ is also performed as an intermediate stage.
We propose in this paper two main strategies to embed the whole graph for the controversy detection needs. 
These strategies rely on hierarchical representations of graphs, convolutional network, and attention-based graph representation.

\subsubsection{Hierarchical graph representation learning-based strategy.} 
This strategy exploit hierarchical structure that may exist in the user graph structure.
Thus, in the whole graph encoding process, graph information are aggregated across the edges iteratively and in a hierarchical way.
 We rely on the \textsc{diffpool} \cite{Ying2018Diffpool} approach which encodes the whole graph by stacking several pooling layers. 
 Each pooling layer is composed of two distinct \textsc{gnn}: one, called $\textsc{gnn}_{embed}$, learns user nodes embeddings $H$, and the other, called $\textsc{gnn}_{pooling}$, learns an assignment matrix $S$ that indicates which user nodes are assigned to which cluster. The matrix $S$ is used to coarsen the graph.


 
 As depicted in  Figure~\ref{fig:diffpool_layer}, the functioning of the pooling layer at level (l)  is described as follows:
 
  
  \begin{figure*}[!htbp]
	\centering
		\includegraphics[width=1\textwidth]{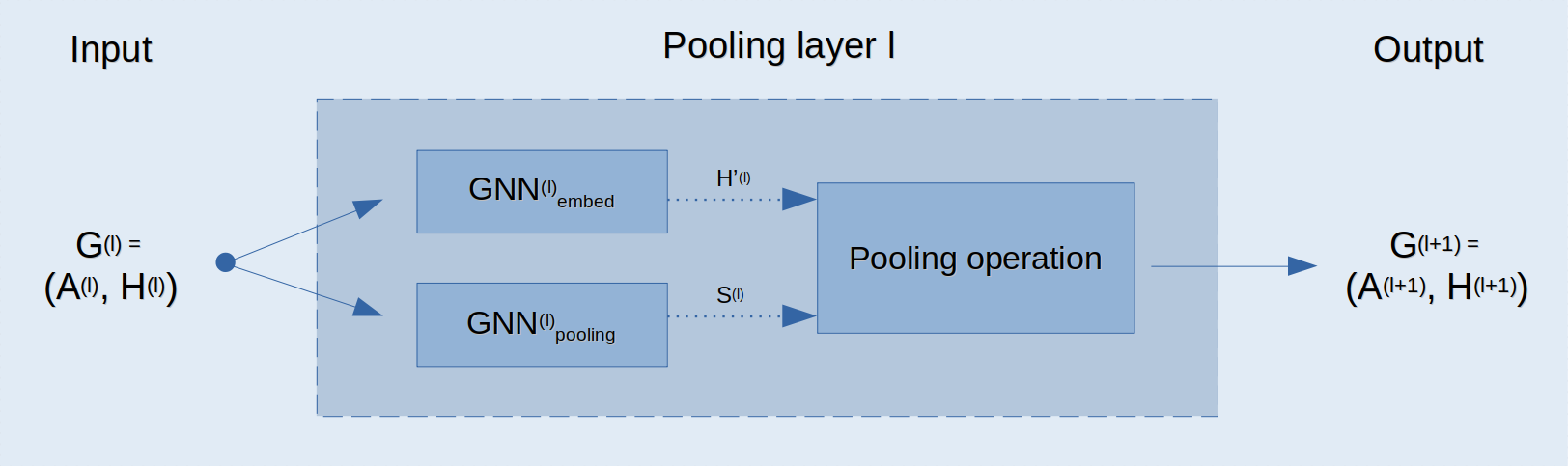}
	\caption{Diffpool-based Pooling layer architecture. $A^{(l)}$ and $H^{(l)}$  represent the adjacency and feature matrices of the input graph at layer ($l$) respectively. $\textsc{gnn}_{embed}$ and $\textsc{gnn}_{pooling}$ are the 2 \textsc{gnn} blocks used to respectively compute node embeddings $H'^{(l)}$ and assignment matrix $S^{(l)}$. The pooling operation block converts the input graph ($A^{(l+1)}$, $H^{(l+1)}$) into a new coarsened graph ($A^{(l+1)}$, $H^{(l+1)}$).}
	\label{fig:diffpool_layer}
\end{figure*}

\begin{enumerate}
    \item \textit{Node embedding generation}. We first apply the $\textsc{gnn}_{embed}^{(l)}$ to the graph obtained at level $(l)$ represented by its adjacency matrix $A^{(l)}$, and its node features matrix $H^{(l)}$. As described in equation~\ref{nodeembeddingequation}, the result is an intermediate node embeddings $H'^{(l)} \in \mathbb{R}^{m \times d'}$, with $m$ number of nodes of the initial graph of the layer, and $d'$ the new dimensional features.
        \begin{equation}
           H'^{(l)} = \textsc{gnn}_{embed}^{(l)} \bigg( A^{(l)}, H^{(l)} \bigg)
           \label{nodeembeddingequation}
        \end{equation}

    \item \textit{Matrix cluster assignment learning}. We then use the $\textsc{gnn}_{pooling}^{(l)}$ to learn a new assignment matrix $S^{(l)}$ to indicate which nodes of the graph at layer $(l)$ will be clustered together to form a new coarser node at layer $(l)$. The matrix assignment is represented by the equation~\ref{matrixassignmentequation}
    \begin{equation}
           S^{(l)} = \textsc{gnn}_{pooling}^{(l)} \bigg( A^{(l)} , H^{(l)} \bigg)
           \label{matrixassignmentequation}
        \end{equation}
    
    \item \textit{Nodes and features pooling}. We finally aggregate nodes belonging to the same cluster and their features from $H'^{(l)}$ using the assignment matrix $S^{(l)}$ to output a new coarser graph represented by its adjacency matrix $A^{(l+1)}$ and features matrix $H^{(l+1)}$.
    This pooling operation is done as follows:
       
        \begin{equation}
           A^{(l+1)} = S^{(l)^{T}} A^{(l)} S^{(l)}
           \label{adjacencymatrix}
        \end{equation}
        \begin{equation}
           H^{(l+1)} = S^{(l)^{T}} H'^{(l)}
           \label{nodefeaturematrix}
        \end{equation}
\end{enumerate}
We can notice that the number of nodes is decreasing at each new layer (l). 
 At the first layer (\(l\)=0), $A^0$ and $H^0$   correspond to the   adjacency matrix  \(A \) and the feature matrix \(X \)  of the initial user graph respectively.  
The last layer L is a single cluster node, and represents the final vector embedding  $z_{G}$ of the whole graph.

In our work, only one kind of \textsc{gnn} is used: Graph Convolutional Networks (\textsc{gcn}) \cite{Kipf2017GCN}.

\subsubsection{Attention mechanism-based user pooling strategy.}  This strategy is based on attention-based node embedding and allows each user node to judge which user neighbors are more important than the others during the node embedding process, according to the structure and features of the graph. Once node embeddings are generated, they are then aggregated to produce the node embedding of the whole graph.
The attention mechanism (\textsc{gat}) with cardinality preservation~\cite{Zhang2020GATGC} is used
to differentiate user neighbors by assigning them different scores. 
This attention mechanism is similar to the Transformers block used in BERT \cite{Devlin2018BERT} for language modelling.

Let's \(\tilde{\mathcal{N}}_{(u_{i})}\) be the multi-set of first-order neighbors of node $u_i$, including $u_i$ itself. This second strategy is described as follows:
\begin{enumerate}
     \item \textit{Neighbors attention score}. For each node \(u_i \in \mathcal{U}\), and each user neighbor \(u_j \in \tilde{\mathcal{N}}_{(u_{i})}\), we first compute the attention score \( e_{u_{i}u_{j}} \) by using an attention function \( a \) on transformed features represented by the matrix \( W^{(l)} \) of the current layer (\(l\)) for both nodes, as described in equation \ref{eq:attention_score}: 
     
     \begin{equation}\label{eq:attention_score}
    e^{(l)}_{u_{i}u_{j}} = a \bigg( \textbf{W}^{(l)} h^{(l)}_{u_{i}}, \textbf{W}^{(l)} h^{(l)}_{u_{j}} \bigg)
     \end{equation}

    \item \textit{Attention scores normalization}. We then normalize scores using a softmax function to get a probability distribution of each score.
    \begin{equation}
        \alpha^{(l)}_{u_{i}u_{j}} = softmax(e^{(l)}_{u_{i}u_{j}}) = \frac{\exp(e^{(l)}_{u_{i}u_{j}})}{\sum_{u_{k} \in \tilde{\mathcal{N}}_{(u_{i})}} \exp(e^{(l)}_{u_{i}u_{k}})}
    \end{equation}

    \item \textit{User node embedding}. The normalized scores are then used to compute the new user node representation \( h^{(l+1)}_{u_{i}} \)
    \begin{equation}\label{simple_head}
        h^{(l+1)}_{u_{i}} = \sigma \Bigg( \sum_{u_{j} \in \tilde{\mathcal{N}}_{(u_{i})}} \alpha^{(l)}_{u_{i}u_{j}} \ \textbf{W}^{(l)} h^{(l)}_{u_{j}} \Bigg)
    \end{equation}
    with \( \sigma \) a non-linear activation function, \(h^{(l+1)}_{u_{i}} \in \mathbb{R}^{n \times d} \) with \(d\) feature dimension of the layer. 
    Note that the cardinality preservation allows in~\ref{simple_head} to scale the result before the use of the activation function \( \sigma \).
    The final node representation \( h_{u_{i}} \) corresponds to the output of the last layer \( h^{(L)}_{u_{i}} \).

    \item \textit{Graph embedding}. Finally, we compute the final graph embedding \( z_G \) by applying the READOUT function. 
    A  simple graph-level pooling function is used: we sum up each node representation at each iteration layer, and then concatenate them as shown in equation~\ref{readout}.
    \begin{equation}
          z_G =  \mathlarger{\mathlarger{\mathlarger{\parallel}}}^{(L)}_{l=0} \bigg( \readout\Big(\{h^{(l)}_{u_{i}} \Big| u_{i} \in U \}\Big) \bigg)
          \label{readout}
    \end{equation}
\end{enumerate}
This attention mechanism presents multiple advantages, in addition to state-of-the-art results in many benchmark graph classification tasks and interpretability. It allows the use of fixed number of parameters,  and therefore does not depend on the graph size. It also presents transductive and inductive capabilities.


\subsection{Graph Classification} 

The graph classification stage aims to classify the post represented by its graph embedding as controversial or non-controversial. 
To do so, we simply rely on a classic multi-layer perceptron classifier with the vector \( z_G \) as input. 
A Softmax activation on the output layer of dimension 2 is used.


\section{Experimental Evaluation}\label{sec:experiment} 

\subsection{Dataset} 
We evaluated the performance of our approach using a real-world Reddit dataset, in English, released by Hessel and Lee~\cite{Hessel2019Controversy}.
The same dataset is also covered by Zhong et al.~\cite{Zhong2020ControversyGCN}.
The collected data covers a period from 2007 to February 2014.
It contains 6 specific online channels (also called subreddit): \textit{AskMen} (\textsc{am}), \textit{AskWomen} (\textsc{aw}), \textit{Fitness} (\textsc{fn}), \textit{LifeProTips} (\textsc{lt}), \textit{personnalfinance} (\textsc{pf}), \textit{relationships} (\textsc{rs}). 
On Reddit, each user can comment on a post (threads) which is related to a specific topic (subreddit).
Each subreddit contains a set of posts. Metadata and a tree-comment structure are associated to each post.
Finally, only posts with a total of at least 30 comments are kept, assuming that less than 30 comments are not enough to build a significant graph.
Each post is automatically labelled controversial or not controversial, depending on various post meta-data~\cite{Hessel2019Controversy}, among them the ratio between up-votes and down-votes~\footnote{Up-vote and down-vote indicate agreement and disagreement on the post.}.
We first separate our data according to the 6 subreddits.
For each subreddit $s$, we create a set of user graphs \( \mathcal{G}_{s} \), one graph per post, each set having at least 1000 posts. 
We then define \( \mathcal{G}_{s,train} \) and \( \mathcal{G}_{s,val} \) as our train and validation graph set respectively, all equally balanced between controversial and non-controversial posts. Considering all aspects and few more experiments, we only evaluate our approach on the same validation set.
The accuracy metric is used to compare the performance of our approach to some existing ones as the dataset is equally balanced.
We separately train the NLP model for user texts representation learning and the \textsc{gnn} for information structure learning.

\subsection{Baseline} 
We compared our approach with the following representative
works on controversy detection using the same Reddit dataset.
Note that those methods perform a k-fold to evaluate their performance, using average accuracy as their metric.
\begin{itemize}
\item (POST \textsc{(Text+Time)})~\cite{Hessel2019Controversy}. It  only focuses on the posts content. It uses language modelling based on BERT~\cite{Devlin2018BERT} and extra-features based on the post timestamp of the post.
\item ({\textsc{C-\{Text\_Rate\_Tree\} + Post}})~\cite{Hessel2019Controversy}. It is based on a simple binary classifier. 
Textual embeddings of a post are combined with structural features of the comment-tree (average representation of text comments, depth of the tree, etc.) of the post and are used as input of the classifier. We compare post with comment during the first hour and the first three hours.
\item (\textsc{DTPC-GCN})~\cite{Zhong2020ControversyGCN}. It is based on a Disentangled Topic-Post-Comment Graph Convolutional Network. Controversial posts are identified by using \textsc{gcn} model and by learning features depending on the respective subreddit post.
\end{itemize}

\subsection{First experiment: Controversy detection based on structural information} 

We implemented our \textsc{gnn}-based controversy detection approach in Pytorch. 
The hierarchical graph representation learning is based on \textsc{diffpool}~\cite{Ying2018Diffpool} and \textsc{gcn}~\cite{Kipf2017GCN}. We refer to it as \textbf{HRL-GCN} (Hierarchical Representation Learning based on \textsc{gcn}).
We also test our strategy by using one and two pooling layers, respectively. For each pooling layer, a 3-layer \textsc{gcn} is used.
We rely on the same loss and optimizer functions used in \textsc{diffpool} experimentation~\cite{Ying2018Diffpool}.
The attention-based node learning is based on \textsc{gat}~\cite{Velickovic2018GAT}, using \textsc{gat-gc} method, with the same hyperparameters used in~\cite{Zhang2020GATGC} . We refer to it as \textbf{ARL-GAT}. We also test this strategy with two different node aggregators to compute the whole graph embedding, namely \textsc{mean} and \textsc{sum} pooling functions.
Both \textsc{gnn}-based strategies are trained with a learning rate at 0.01, a batch size at 32 during 100 epochs. Table \ref{tab:stats_dataset} shows statistics on the different datasets.

\begin{table}
    \centering
    \caption{Statistics on the 6 real-world balanced Reddit datasets.}  \label{tab:stats_dataset}
    
    \begin{tabular}{|l|l|l|l|l|l|l|}
    \hline
     & \ \textsc{am} & \ \textsc{aw} & \ \textsc{fn} & \ \textsc{ls} & \ \textsc{pf} & \ \textsc{rs} \\
    \hline
    Number of posts & \ 3305 \ & \ 2969 \ & \ 3934 \ & \ 1573 \ & \ 1004 \ & \ 2248 \ \\
    \hline
    Average number of user by post \ & \ 72  & \ 67 & \ 76 & \ 79 & \ 47 & \ 48 \\
    \hline
    Average number of comment by post \ & \ 144  & \ 141 & \ 159 & \ 132 & \ 95 & \ 98 \\
    \hline
    Average number of words by comment \ & \ 41  & \ 42 & \ 34 & \ 28 & \ 52 & \ 61 \\
    \hline
    \textcolor{black}{Ratio of comments with tokens $\geq$ 256} & \ 2.68  & \ 2.64 & \ 1.61 & \ 1.03 & \ 4.1 & \ 6.17 \\
    \hline
    \end{tabular}
\end{table}

First experiments are performed without text representation to underline the importance of structural interaction between users in controversial discussion.
Table~\ref{tab:results_gnn} reports the accuracy results 
where the first four lines correspond to the baseline and the last four lines correspond to our experiments results.

\begin{table}
    \centering
    \caption{Performance comparison of our \textsc{gnn}-based controversy detection with baseline. Performance is evaluated using accuracy of the validation set.}  \label{tab:results_gnn}
    
    \begin{tabular}{|l|l|l|l|l|l|l|}
    \hline
     & \ \textsc{am} & \ \textsc{aw} & \ \textsc{fn} & \ \textsc{ls} & \ \textsc{pf} & \ \textsc{rs} \\
    \hline
    POST \textsc{(Text+Time)} & \ 68.1 \ & \ 65.4 \ & \ 65.5 \ & \ 66.2 \ & \ 66.5 \ & \ 69.3 \ \\
    \hline
    \textsc{DTPC-GCN} & \multicolumn{6}{c|}{67.6} \\ 
    \cline{2-7}
    POST + \textsc{C-\{Text\_Rate\_Tree\}} $<$ 1 hour \ & \ 71.1  & \ 70 & \ 68.1 & \ 67.9 & \ 66.1 & \ 65.5 \\
    POST + \textsc{C-\{Text\_Rate\_Tree\}} $<$ 3 hours \ & \ \textbf{74.3} & \ 72.3 & \ 70.5 & \ \textbf{71.8} & \ \textbf{69.3} & \ \textbf{67.8} \\
    \hline
    \textsc{ARL-GAT} (\textsc{mean}-aggr) & \ 65.7 & \ 69.2  & \ \underline{\textbf{72.4}} & \ 58.4 & \ 53.7 & \ 62.9 \\
    \textsc{ARL-GAT} (\textsc{sum}-aggr) & \ 67.5 & \ 71 & \ 72.2 & \ 67 & \ 63.7 & \ 51.8 \\
    \textsc{HRL-GCN} (pool=2) & \ 69 & \ 72.2 & \ 71.7 & \ \underline{68.3} & \ 65.7 & \ 63.6 \\
    \textsc{HRL-GCN} (pool=1) & \ \underline{69.6} & \ \underline{\textbf{74.6}} & \ 72.2 & \ 67.9 & \ \underline{68.2} & \ \underline{66.7} \\
    \hline
    \end{tabular}
\end{table}


As shown in Table \ref{tab:results_gnn}, our hierarchical approach (\textsc{HRL-GCN}) gets the best results among our experiments, with a weighted average accuracy at $70.6$ using only one pooling layer. Our Attention-based approach \textsc{ARL-GAT} reaches $66.8$ and $66.2$ with the \textsc{sum} and \textsc{mean} aggregator, respectively.
\textsc{HRL-GCN} beats the \textsc{DPTC-GCN}~\cite{Zhong2020ControversyGCN} method and the hybrid method proposed by Hessel and Lee~\cite{Hessel2019Controversy} with comments of the first hour for almost every dataset.
Our proposed method (\textsc{HRL-GCN}, pool=1) gets around state-of-the-art results on several datasets, even going up to $74.6$ accuracy in the \textsc{aw} dataset, beating results in \textsc{C-\{Text\_Rate\_Tree\} + Post} with comments of more than the first three hours.
As the \textsc{am} dataset is the biggest dataset, it could mean that our approach generalizes better when data are abundant, and less when data are sparce.
As explained in~\cite{Hessel2019Controversy}, not enough comments are available for each dataset of the baseline. Indeed, when the subreddit \textit{AskMen} (\textsc{am}) has in average 10 comments after 45min, \textit{Relationships} (\textsc{rs}) does not even have those after 3 hours.
With more available comments, we might have a better embedding representation of our graph, which could lead to better performance.

Table \ref{tab:results_gnn} shows that our attention-based approach \textsc{ARL-GAT}, combined with the \textsc{mean}-aggregator, performs well in the three first datasets, beating our best baseline method on \textsc{fn}, with an accuracy of $72.4$ on the validation set. On the other hand, it underperforms on the other three, falling to $53.7$ on \textsc{pf}.
\textsc{pf} and \textsc{rs} already have low results on our baseline, which means that the data is difficult to understand. This could also be explained by the fact that those 3 subreddits have the least average number of comments (as shown in Table \ref{tab:stats_dataset}), and therefore each user node has less neighbors. Attention scores are in fact less useful in these cases. In general, higher average degree of nodes 
could lead to better performance.


\subsection{Second experiment: textual content and structural information based on controversy detection}

We conducted a second experiment to study the impact of adding textual node features to our \textsc{gnn}-based architecture.
Instead of considering all options of our \textsc{gnn}-based architecture shown in Table~\ref{tab:results_gnn}, we only considered our hierarchical representations strategy \textsc{HRL-GCN}, with one pooling layer, as it realises the best accuracy scores.

Text features of comments and posts are extracted using different language model based on BERT, and are aggregated by user to be used as the initial features of our user nodes.

We use different models to extract those features:
\begin{itemize}
    \item \textsc{PT} model. It only uses the pre-trained features to get the message representation. The last layer (768 dimensions) is outputted as our text embeddings.
   
   \item \textsc{FT\_itself} model. 
    We fine-tune \cite{Sun2020BERTFineTuning} a BERT model using comments and posts of our train set \( \mathcal{G}_{s,train} \), with an extra-layer of 64 neurons on top, in addition to the classifier layer. We label each comment with the controversy label of its respective post.
    Note that each subreddit is fine-tuned separately as we suppose that different communities express themselves differently, and texts can be interpreted differently.
    
    \item \textsc{FT\_sentiment} model. We fine-tune a BERT model using sentiment analysis with another Reddit dataset of comments (hosted on kaggle.com),
    labeled as negative, positive or neutral. Indeed, we suppose here that sentiments can outline users' behavior on controversial posts.
    
\end{itemize}
In all cases, we use the 'base-bert-uncased' version (with its corresponding tokenizer), with 12 transformer layers and 110 millions parameters. For time and memory performance, we only use 256 tokens max per text (instead of 512, as Table \ref{tab:stats_dataset} shows that in average, less than 3\% of the messages are represented by more than 256 tokens). For fine-tuning models, we use the same hyperparameters used in \cite{Sun2020BERTFineTuning}.
Table~\ref{tab:results_gnn_text} shows the new accuracy scores obtained by incorporating text features in our \textsc{HRL-GCN} strategy.

\begin{table}
    \centering
    \caption{Performance of our best \textsc{gnn} approach enriched with different user text embeddings as initial node features.}\label{tab:results_gnn_text}
    \begin{tabular}{|l|l|l|l|l|l|l|}
    \hline
     & \ \textsc{am} & \ \textsc{aw} & \ \textsc{fn} & \ \textsc{ls} & \ \textsc{pf} & \ \textsc{rs} \\
    \hline
    \ \textsc{HRL-GCN} (pool=1) \ \ \ \ & \ 69.6 \ & \ \underline{\textbf{74.6}} \ & \ \underline{72.2} \ & \ 67.9 \ & \ 68.2 \ & \ \underline{66.7} \ \\
    \hline
    \ \ \ \ + \textsc{FT\_sentiment} \ & \ 69.1 \ & \ 72.9 \ & \ 70.5 \ & \ \underline{68.6} \ & \ 66.7 \ & \ 64 \ \\
    \ \ \ \ + \textsc{FT\_itself} \ & \ 67.3 \ & \ 73.9 \ & \ 71.8 \ & \ 68.3 \ & \ \underline{70.6} \ & \ 63.8 \ \\
    \ \ \ \ + \textsc{PT} \ & \ \underline{70.8} \ & \ 73.7 \ & \ 71 \ & \ 65.4 \ & \ \underline{70.6} \ & \ 64.7 \ \\
    \hline
    \end{tabular}
\end{table}


For three of the six datasets, adding textual features improve controversy detection results.
Our \textsc{HRL-GCN} strategy combined with the pre-trained (\textsc{PT}) BERT features gets better results when using \textsc{am} and \textsc{pf} datasets, with  $70.8$ and $70.6$ accuracy result respectively.
Adding sentiment features from \textsc{FT\_sentiment} allows us to increase accuracy from $67.9$ to $68.6$  on \textsc{ls}.
Even if the content has interesting features for controversy detection, it remains brittle and community specific, which means that textual features can be more impactful in some subreddits than others. For instance, sentiments about controversial topic can be more meaningful in subreddit \textit{LifeProTips} (\textsc{lt}) than in more personal subreddits, like \textit{Relationships} (\textsc{rs}).
The complexity of the data and the fact that datasets might be too small compared to the number of features (which goes up to $768$ when using \textsc{PT} model) could also explain why our \textsc{gnn}-based models overfit on some datasets, and therefore does not improve accuracy results.


  \begin{figure*}[!htbp]
	\centering
		\includegraphics[width=0.55\textwidth]{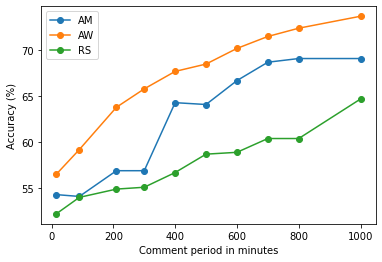}
	\caption{Impact of comments availability on controversy detection performance.}
	\label{fig:plot_accuracy_timebased}
\end{figure*}

Figure \ref{fig:plot_accuracy_timebased} shows the importance of comments availability.
It reports accuracy results evolution over time (minutes) of three datasets when using our best \textsc{HRL-GCN} strategy combined with text features from PT. 
It clearly shows that the more available comments we have, the easier it is to detect controversy.

\section{Conclusion} 

We presented an automatic controversy detection method on social media, based on \textsc{gnn} techniques.
We considered this detection as a classification task and first exploited the 
structural information that characterizes user interactions by
defining two strategies of graph embedding. 
The first strategy exploits hierarchical structure that may exist in the user graph.
The second strategy allows each user to select its neighbors in the embedding nodes process.
We also improved the graph embedding by incorporating textual content features 
computed from BERT model.
Experimental evaluation shows promising results, even beating our baseline in several datasets~\footnote{\textcolor{black}{This work was supported by grants from Janssen Horizon endowment fund}}. However, our current Reddit dataset shows its limits, as a post has usually few comments, which prevents our \textsc{gnn}-based model from getting a better graph representation for controversy detection. The use of a different platform, such as Twitter, which provides more data per topics, \textcolor{black}{or Wikipedia}, could be an interesting lead to follow. 
In terms of future work, we would like to examine the appropriateness
of other \textsc{gnn} techniques for controversy detection. For instance, it could be interesting to study the impact, in terms of
performance improvements, of using \textsc{gnn} architectures that take into account nodes properties,
\textcolor{black}{mixing then structural, textual and user information.}
\textcolor{black}{compare results from different social media at the same time could also help to have a better understanding of the subject covered. Quantifying controversy using our approach is also an interesting perspective.}


\bibliographystyle{splncs04}
\bibliography{biblio_paper.bib}

\end{document}